\documentclass[10pt,letterpaper]{article}

\usepackage{cogsci}
\usepackage{apacite}

\usepackage{pslatex}

\usepackage{amsmath}
\usepackage{caption}
\usepackage{xcolor}
\usepackage{graphicx}
\usepackage{multirow}
\usepackage{subcaption}
\usepackage{tabularx}
\usepackage{booktabs}

\usepackage{times}
\usepackage{url}

\usepackage{authblk}

\newcommand\ie{\emph{i.e.}}
\newcommand\eg{\emph{e.g.}}

\newcommand{\Example}[1]{Ex.~\ref{#1}}
\newcommand{\Figure}[1]{Figure~\ref{#1}}
\newcommand{\Equation}[1]{Eqn.~(\ref{#1})}

\newcommand\rep{\mathrm{v}}
\newcommand\simf{\mathrm{sim}}

\graphicspath{{plots/}{}}

\title{Calculating Probabilities Simplifies Word Learning}

\author[1]{\bf Aida Nematzadeh\thanks{nematzadeh@berkeley.edu}}
\author[2]{\bf Barend Beekhuizen\thanks{barend@cs.toronto.edu}}
\author[2]{\bf Shanshan Huang\thanks{sunny.huang@mail.utoronto.ca}}
\author[2]{\bf Suzanne Stevenson\thanks{suzanne@cs.toronto.edu}}
\affil[1]{Dept.\ of Psychology, Univ.\ of Calif., Berkeley}
\affil[2]{Dept.\ of Computer Science, University of Toronto}

\begin{document}

\maketitle

\begin{abstract}
Children can use the statistical regularities of their environment to learn
word meanings, a mechanism known as cross-situational learning. We take a
computational approach to investigate how the information present during each
observation in a cross-situational framework can affect the overall acquisition
of word meanings. We do so by formulating various in-the-moment learning
mechanisms that are sensitive to different statistics of the environment, such
as counts and conditional probabilities. Each mechanism introduces a unique
source of competition or mutual exclusivity bias to the model; the mechanism
that maximally uses the model's knowledge of word meanings performs the best.
Moreover, the gap between this mechanism and others is amplified in more
challenging learning scenarios, such as learning from few examples.  
\end{abstract}

\section{Introduction}
\label{sec:intro}

How do people acquire the meanings of words as they begin to learn a language?
A well-supported proposal is \textit{cross-situational learning}
\cite<\eg,>{pinker.1989}, which suggests that people are sensitive to the
regularities that repeat in different situations, and use such evidence to
identify the commonalities, from which they can infer word meanings.  As an
example, when a child hears \textit{what a cute kitty}, \textit{be nice to the
kitty}, etc., she/he could infer that the word \textit{kitty} refers to the
common referent in all these situations, \ie, a cat. Recent word learning
experiments confirm that both adults and infants keep track of
cross-situational statistics across learning trials, and infer the correct
word--meaning mappings even in highly ambiguous conditions 
\cite<\eg,>{yu.smith.2007, smith.yu.2008, yurovsky.etal.2014}.

Despite empirical evidence for statistical cross-situational learning, the
exact mechanisms in play are still not fully understood.  In this paper, we
focus on the first step of a cross-situational framework -- the learning that
occurs on each observation of a word, which we call \textit{in-the-moment
learning}.  Given the words in an utterance and their potential meanings in the
accompanying situation, there are many possible ways to associate words and
meanings, but only some of these associations are correct.  We refer to these
in-the-moment associations of words and meanings as \textit{alignments}, and
consider different strategies for assessing the strength of these alignments,
drawing on the evolving knowledge of word meanings.  We note that previous
research has considered ``hard'' (or binary) in-the-moment learning strategies,
where an alignment is either considered by the learner or not
\cite<\eg,>{trueswell.etal.2013}; we instead examine ``soft'' strategies where
alignments have strengths between zero and one.

We formulate various in-the-moment learning mechanisms that introduce different
kinds of competition -- i.e., the way in which the strength of a word--meaning
alignment depends on and interacts with other possible alignments.
Each mechanism corresponds to certain statistics of the word learning input,
such as the weighted frequency of word--meaning pairs or their conditional
probabilities.
We show that the different types of competition % and statistics considered
during the in-the-moment learning lead to various kinds of mutual exclusivity
behaviours.
Mutual exclusivity has been proposed as an explicit bias, in which children
assume each word has a single meaning \cite<\eg,>{markman.1987,
markman.wachtel.1988}.
Here, mutual exclusivity of words and/or meanings arises from competition in a
way that focuses learning.

We take a computational modeling approach to investigate the effectiveness of
these mechanisms in overall acquisition of word meanings in various long-term
word learning scenarios.
Using a computational model enables us to explore the impact of different
learning mechanisms in a variety of conditions, and to examine the role of one
factor (\eg, frequency) while controlling for another one (\eg, utterance
length).
We find that the mechanism that maximizes the use of the accumulated knowledge
of learned meanings performs the best. Interestingly, the performance gap
between this mechanism and others is most significant in more difficult
learning conditions, such as learning of low frequency words given long
utterances. This shows that using conditional probabilities (as opposed to
counts) and introducing competition (leading to a mutual exclusivity bias)
improves overall word learning and might be necessary to guide learning in the
presence of ambiguity or little data.

\section{A Cross-situational Word Learning Framework} \label{seq:framework}

There has been an increased interest in the last decade in developing
computational models as tools to study word learning in people.  Of particular
interest are cross-situational learners that are incremental
\cite<\eg,>{siskind.1996, fazly.etal.2010.csj, kachergis.etal.2012.psyrev},
which is necessary in studying developmental learning patterns.
Notably, the model of \citeauthor{fazly.etal.2010.csj}
\citeyear{fazly.etal.2008.cogsci,fazly.etal.2010.csj} (henceforth FAS) is the
first probabilistic model that robustly predicts a range of observed behavior
in child word learning. Moreover, this model has been adopted and extended by a
series of successive work \cite<\eg,>{nematzadeh.etal.2012.cmcl,
grant.etal.2016}, demonstrating its robustness in accounting for empirical
data.  We adopt the FAS word learning framework to examine various
in-the-moment learning mechanisms.

\subsection{The FAS Model}

\noindent\textbf{Word learning input and output.} The model's input is a
sequence of utterance--scene pairs simulating what the child hears and
perceives, respectively. Each utterance is a set of words (ignoring their
order), and the corresponding scene is a set of semantic features that
represents possible meanings of words in the utterance.  Word meanings are
represented by multiple features, which exposes the model to naturalistic
commonalities among the words.

\vspace{-.4cm}
\begin{eqnarray}
{\small
\begin{tabular}{l}
{\bf Utterance:} $\{$ \emph{Joel}, \emph{eats} $\}$ \\
{\bf Scene:} $\{$ \textsc{person}, \textsc{joel}, \textsc{act}, \textsc{consume}, ... $\}$ \\
\end{tabular}}
\label{US-features}
\end{eqnarray}

\vspace{-.2cm}

\noindent
The output of the model, at each step in learning, is the current
representation of the meaning of each word $w$ as a probability distribution,
$p(\cdot|w)$, over all possible semantic features $f$ that the model has
observed in the input scenes.

\noindent\textbf{The word learning problem.} Given a corpus of utterance--scene
pairs, the goal of the model is to learn the meaning probability distribution,
$p(\cdot|w)$, for all words $w$.  Prior to receiving any input, all features
$f$ are equally likely for a word.  As the model processes each input pair, the
probability is adjusted to reflect the cross-situational evidence in the
corpus, in two steps: (a) in-the-moment learning on this input pair and (b)
update of the word meaning probabilities using the accumulated evidence over
all inputs.

\vspace{.1cm} \noindent\textbf{In-the-moment learning.} Given an utterance and
a scene, which features in the scene are part of a word's meaning?  There are
different possible ways to determine whether a semantic feature is associated
with a word in the input pair, and the corresponding strength of that
association. FAS assumes that each feature $f$ in scene $S_t$ at time $t$,
independently of the other features, is \textit{aligned} to all the words $w$
in the utterance $U_t$ with a particular strength (see \Figure{fig:align-FAS}):

\vspace{-.4cm}
\begin{eqnarray}
\mathit{a_{t}(w|f) = \displaystyle\frac{p_t(f|w)}{\displaystyle\sum_{w' \in \mathrm{U}_t}{p_t}(f|w')}}
\label{eq:FASalign}
\end{eqnarray}
\vspace{-.4cm}

\noindent
The alignment strength between a feature $f$ and word $w$ depends on the
current probability that $f$ is part of the meaning of $w$ -- i.e., $p_t(f|w)$
-- as well as the probabilities that $f$ is part of the meaning of other words
in the utterance (the denominator above).

In this way, \Equation{eq:FASalign} has words in the utterance ``compete'' to
be associated with a given feature: a higher alignment strength of one word
with a feature necessarily results in a lower alignment strength for other
words with that feature. This can be interpreted as a directional
\textit{mutual exclusivity bias}: the alignment formulation limits the number
of words a feature can be strongly associated with, but does not directly limit
the number of features a word can be associated with.

\vspace{.1cm}
\noindent\textbf{Updating the word meanings.} How is the information learned
from an input pair incorporated into a learner's long-term knowledge of word
meanings? The learner incrementally accumulates the alignment strengths between
each $w$ and $f$ in an overall association score $assoc(w,\,f)$, which is
updated at each time $t$ that $w$ and $f$ co-occur in an input pair:

\vspace{-.4cm}
\begin{eqnarray}
\mathrm{assoc}_{t}(w,\,f) = \mathrm{assoc}_{t-1}(w,\,f) + a_{t}(f|w)
\label{eq:FASassoc}
\end{eqnarray}
\vspace{-.4cm}

\noindent
where $\mathrm{assoc}_{t-1}(w,\,f)=0$ if $w$ and $f$ have not co-occurred prior to $t$.

After updating the association scores, the meaning probability $p(\cdot|w)$ of
each word $w$ in the current input is adjusted using a smoothed version of this
formula:

\vspace{-.4cm}
\begin{eqnarray}
\mathit{p_{t+1}(f|w) = \displaystyle\frac{\mathrm{assoc}_{t}(f,\,w)}
          {\displaystyle\sum_{f_j \in \mathcal{M}}{\mathrm{assoc}_{t}(f_j,\,w)}}}
\label{eq:FASmprob}
 \end{eqnarray}
\vspace{-.4cm}

\noindent where $\mathcal{M}$ is the set of all features observed up to time
$t$.  In \Equation{eq:FASmprob}, the probability of a feature given a word is a
normalization of their association score over all possible features, which
introduces another source of competition, this time, among features for a given
word.  This competition can be thought of as a mutual exclusivity bias in the
reversed direction of the alignment score in \Equation{eq:FASalign}; here a
word can only be strongly associated to a limited number of features.

\subsection{Grouping Sets of Features into Referents}
In FAS, an input scene is the set union of all meaning features for all words
in the corresponding utterance.  This representation lacks information that
would be apparent to a child concerning how groups of features belong to a
single entity or event -- e.g., \textsc{person} and \textsc{joel}, or
\textsc{act} and \textsc{consume} in \Example{US-features}.  However, replacing
the sets of features with a single symbol corresponding to the meaning would
prevent the model from learning semantic similarities among the words
\cite<\eg,>{nematzadeh.etal.2012.cogsci}.  Instead, following
\citeA{alishahi.etal.2012}, we group the semantic features in a scene into
\textit{referents} that correspond to something referred to by a word in the
utterance, as in \Example{US-referents}:\footnote{We use the term
\textit{referent} to denote anything referred to by a word -- an object or
event, or set of semantic properties (e.g., $\{$ \textsc{indefinite},
\textsc{singular} $\}$ for \textit{an}).}

\vspace{-.4cm}
\begin{eqnarray}
{\small
\begin{tabular}{l}
{\bf Utterance} $\{$ \emph{Joel}, \emph{eats}, \emph{an}, \emph{apple} $\}$ \\
{\bf Scene:} $\{$  \textsc{$\{$person, joel$\}$},
                  \textsc{$\{$act, consume, ...$\}$}, \\
                  \textsc{$\{$singular, indefinite, determiner, ... $\}$}, \\ \textsc{$\{$apple, fruit, food, ...$\}$} $\}$ \\
\end{tabular}
}
\label{US-referents}
\end{eqnarray}
\vspace{-.4cm}

A scene is now a set of referents, each of which is a set of semantic features.
In the FAS model, calculation of alignment strength between a word $w$ and
feature $f$ at time $t$ uses the meaning probability $p_t(f|w)$.  Now, aligning
words with \textit{referents} (as in~\ref{US-referents}) requires consideration
of strength of alignment of a word with a \textit{set} of features.  In
calculating alignment strength for a word $w$ and a referent $r$ at time $t$,
we change the FAS model to consider $\simf(\rep_t(w),\, \rep(r))$, the
similarity between the word's current meaning representation and the
representation of the referent, where $\rep(r)$ and $\rep_t(w)$ are vectors in
which the elements are meaning features.  For $\rep_t(w)$, the value for each
component feature $f$ is $p_t(f|w)$.  (I.e, $\rep_t(w)$ is a vector
corresponding to $p_t(\cdot|w)$.)  For $\rep(r)$, the element values are $1$
for features present in the definition of $r$ and $0$ otherwise.  In this way,
alignment strength for word $w$ and referent $r$ is influenced by the strength
of the meaning probabilities $p(f|w)$ for all features $f$ that are part of the
representation of $r$.
In the remainder of the paper, we explore variations in how the alignment
process actually does this, in ways that implement different types of mutual
exclusivity biases.

\subsection{In-the-Moment Learning Mechanisms}

\vspace{.1cm}
\noindent\textbf{Competition in the model.} We observed above that the
alignment strength calculation in \Equation{eq:FASalign} instantiates a form of
mutual exclusivity bias, because words are competing to be strongly associated
with a feature during this in-the-moment learning process.  With the change of
aligning words to referents instead of to features, we have the opportunity to
explore various ways to formulate competition in determining the strength of
alignments.  The three alignment formulations explored here implement (1) no
competition among words or referents, (2) competition of referents for a word
(as in  \citeNP{alishahi.etal.2012}), and (3) competition of words for a
referent (analogous to the competition of words for a feature in FAS).  Each of
these ways of viewing competition implements a different approach to mutual
exclusivity in the model, and we will explore the resulting impact on word
learning in the results.

\begin{figure*}
\centering
\begin{subfigure}{.24\textwidth}
\includegraphics[width = \linewidth,height=25mm]{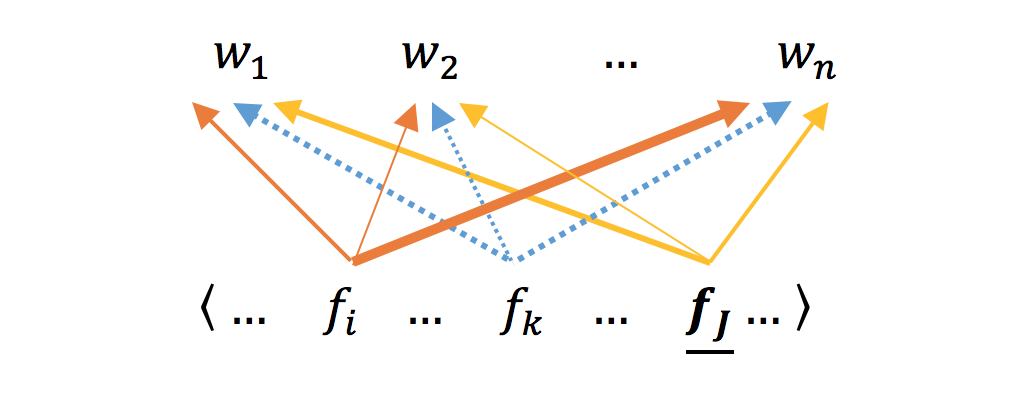}
\caption{FAS.}
\label{fig:align-FAS}
\end{subfigure}
\begin{subfigure}{.24\textwidth}
\includegraphics[width = \linewidth,height=25mm]{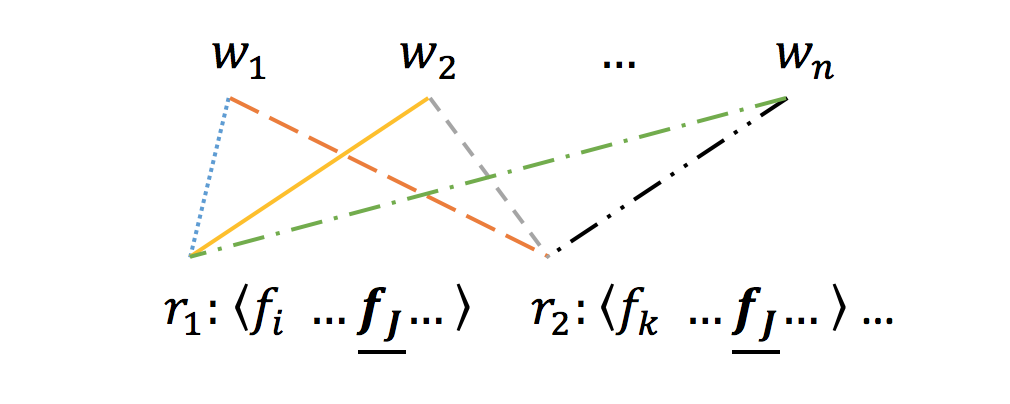}
\caption{No competition.}
\label{fig:align-nocomp}
\end{subfigure}
\begin{subfigure}{.24\textwidth}
\includegraphics[width = \linewidth,height=25mm]{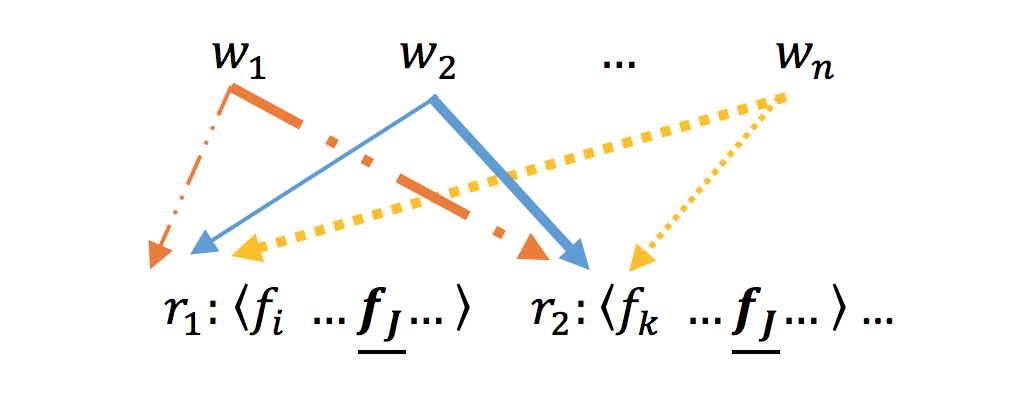}
\caption{Referent competition.}
\label{fig:align-word2refs}
\end{subfigure}
\begin{subfigure}{.24\textwidth}
\includegraphics[width = \linewidth,height=25mm]{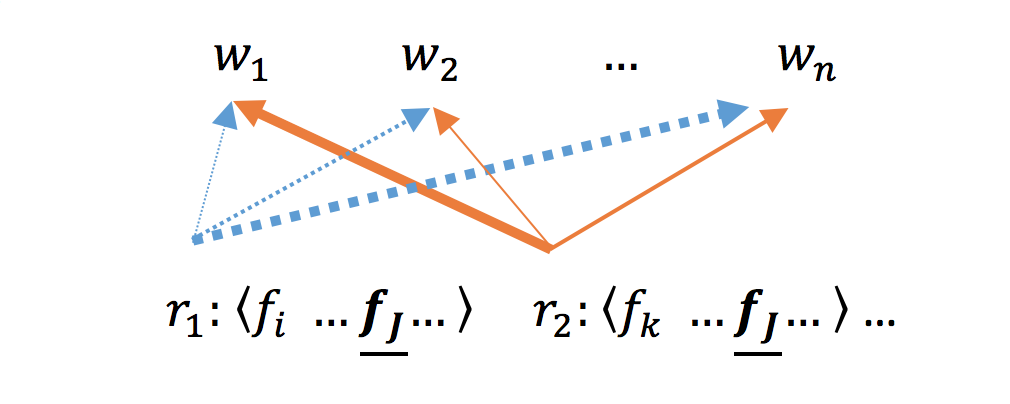}
\caption{Word competition.}
\label{fig:align-ref2words}
\end{subfigure}
\vspace{-0.2cm}
\caption{Types of alignment mechanisms. Lines of the same color/style compete simultaneously.  Thickness indicates varying strength of alignment during a competition.}
\label{fig:align}
\vspace{-0.4cm}
\end{figure*}

\vspace{.1cm}
\noindent\textbf{No competition.} The no-competition mechanism (henceforth,
no-comp) serves as a baseline for comparison to the other two.  It assumes no
mutual exclusivity bias -- all the alignments between words and referents are
calculated independently, and the value of one alignment does not effect any of
the others (see \Figure{fig:align-nocomp}).  We formulate such an alignment
between a word $w$ and a referent $r$ as simply the similarity between
$\rep_t(w)$ and $\rep(r)$ as described above:

\vspace{-.4cm}
\begin{eqnarray}
\mathit{a_{t}(w,\,r) = \simf(\rep_t(w),\, \rep(r))}
\end{eqnarray}
\vspace{-.4cm}

\noindent
This formulation can be seen as a simple weighted count, where each feature
relevant to $r$ (valued $1$ in $\rep(r)$) contributes to the overall alignment
strength proportionally to the model's prior knowledge of its meaning
probability with that word.

\vspace{.1cm}
\noindent\textbf{Referent competition.} Here we adopt the alignment formulation
of \citeA{alishahi.etal.2012}, which we call ``ref-comp'' because referents
compete for alignment with a word.  This mechanism implements a directed mutual
exclusivity bias in which each word has a preference to be strongly associated
with one referent.  In other words, referents in the scene compete for a given
word, while the alignments of words are independent of each other (see
\Figure{fig:align-word2refs}).  This preference can be implemented by
normalizing the $\simf(\rep_t(w),\, \rep(r))$ over all the referents in the
scene:

\vspace{-.4cm}
\begin{eqnarray}
\mathit{a_{t}(r|w) = \displaystyle\frac{\simf(\rep_t(w),\, \rep(r))}
{\displaystyle\sum_{r' \in \mathrm{S}_t}{\simf(\rep_t(w),\, \rep(r'))}}}
\label{eq:w2ralign}
\end{eqnarray}
\vspace{-.4cm}

\noindent
By normalizing the weighted count of $\simf(\rep_t(w),\, \rep(r))$, this
alignment formulation can be interpreted as the conditional probability of $r$
given $w$, rather than a simple count.

\vspace{.1cm}
\noindent\textbf{Word competition.} Here, we consider a competition that is
instead analogous to the competition of words for a feature in FAS;
``word-comp'' is the reverse of ref-comp, because here words compete for a
referent.  This leads to a directed mutual exclusivity bias, but in the
opposite direction to ref-comp.  The word-comp mechanism asserts a preference
for each referent to be strongly associated with a single word, by having words
compete for a referent, while the alignments of referents are independent of
each other  (see \Figure{fig:align-ref2words}).  This bias is formulated by
normalizing the $\simf(\rep_t(w),\, \rep(r))$ over the words in the utterance
(as FAS did):

\vspace{-.4cm}
\begin{eqnarray}
\mathit{a_{t}(w|r) = \displaystyle\frac{\simf(\rep_t(w),\, \rep(r))}
{\displaystyle\sum_{w' \in \mathrm{U}_t}{\simf(\rep_t(w'),\, \rep(r))}}}
\end{eqnarray}
\vspace{-.4cm}

\noindent
This formulation also yields a conditional probability, but here of $w$ given $r$.

\vspace{.1cm}
\noindent\textbf{The association score.} We note one final change to the FAS
model to deal with referents: We must modify \Equation{eq:FASassoc} to keep
track of associations between a word $w$ and all the features of a referent
$r$. Since a feature $f$ can occur in more than one referent in scene $S$,
which can have multiple alignment scores, we use the maximum alignment score of
a referent that contains the feature in updating the feature's association
score:

\vspace{-.4cm}
\begin{eqnarray}
\mathrm{assoc}_{t}(w,\,f) = \mathrm{assoc}_{t-\mathrm{1}}(w,\,f) + \underset{r' \in S : f \in r'}{\max}\mathit{a_{t}(w,\,r')}
\label{eq:new-assoc}
\end{eqnarray}
\vspace{-.4cm}

The meaning probabilities in the model continue to be calculated between
individual features and a word. Recall that the meaning probability
distribution $p(\cdot|w)$, as a conditional probability over semantic features,
enforces a competition among them for the probability mass.

\section{Experiments}
\label{sec:results}

\subsection{Set-up}

The utterances in the input are child-directed speech taken from the Manchester
corpus \cite{theakston.etal.2001} in CHILDES \cite{macwhinney.2000}. To create
the associated scene representations, each word in the corpus is entered into a
gold-standard lexicon with a set of semantic features representing its
gold-standard meaning, following the procedure of
\citeA{fazly.etal.2008.cogsci}. The referents shown in \Example{US-features}
correspond to the gold-standard meanings of each of those words.  (The
word--mapping in the lexicon is only used to generate scenes, and is not seen
by the model.)  The model is trained on $20$K utterance--scene pairs, at which
point behaviour is stable.
 
In the following  experiments, we examine the quality of the individual learned
word representations in two ways: the average acquisition score of all words
observed by the model, and the proportion of observed words that is learned.
The acquisition score of each word $w$ is obtained by comparing the word
meaning representation $\rep(w)$ with a gold standard representation of the
word $\text{gold}(w)$ using cosine similarity:

\vspace{-0.4cm}
\begin{equation}
    \mathbf{acq}(w) = \text{sim}(\rep(w),\text{gold}(w))
    \label{eq:acq-score}
\end{equation}

\noindent
where $\text{gold}(w)$ is a vector over all semantic features, with value $1$
for features part of the gold-standard meaning of $w$ and $0$ otherwise.  An
observed word counts as ``learned'' if its $\mathbf{acq}$ score is higher than
some threshold $\theta$, here set to $0.7$.

\subsection{Results}

\vspace{.1cm}
\noindent\textbf{Overall Learning Patterns}

\begin{figure}
\begin{subfigure}{.49\linewidth}
\includegraphics[width = \linewidth]{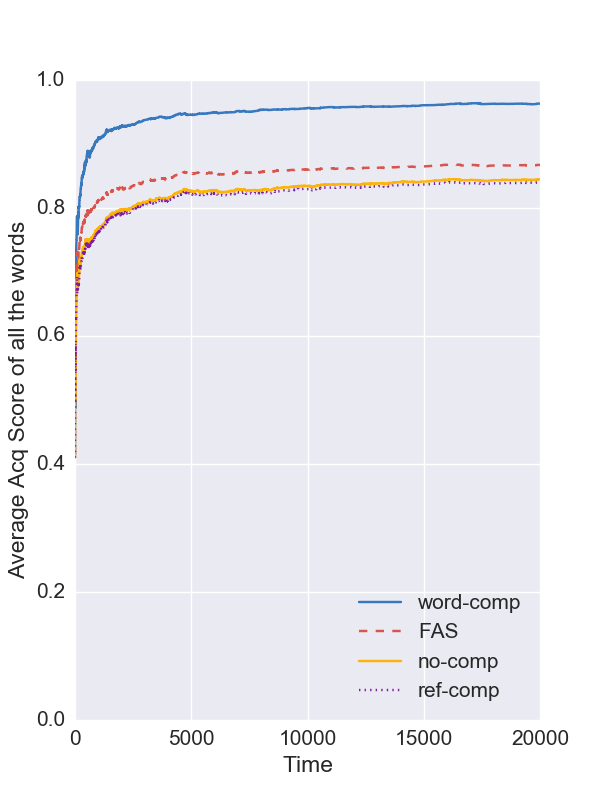}
\caption{Average acquisition score over time.}
\label{fig:res-acq}
\end{subfigure}
\begin{subfigure}{.49\linewidth}
\includegraphics[width = \linewidth]{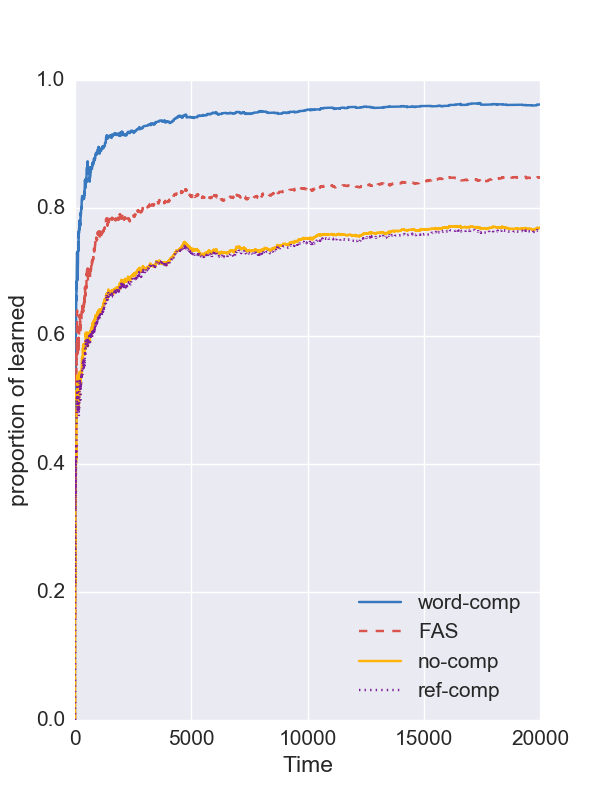}
\caption{Proportion of learned words over time.}
\label{fig:res-proplearned}
\end{subfigure}
\vspace{-0.3cm}
\caption{Developmental plots}
\vspace{-0.4cm}
\end{figure}

\begin{figure*}[h]
\begin{subfigure}{.32\textwidth}
\includegraphics[width = \linewidth]{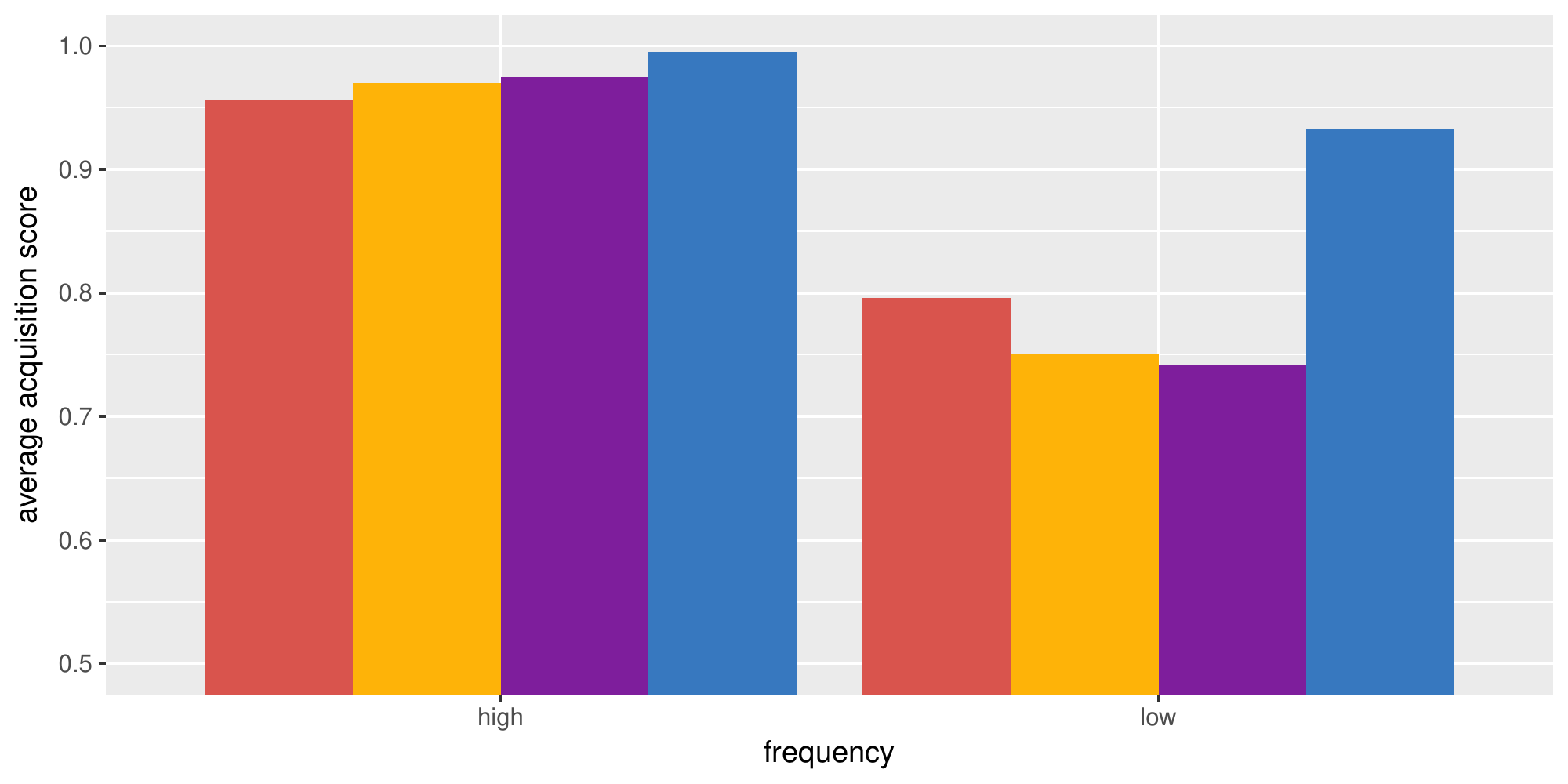}
\caption{Split over word frequency.}
\label{fig:res-acq-freq}
\end{subfigure}
\begin{subfigure}{.32\textwidth}
\includegraphics[width = \linewidth]{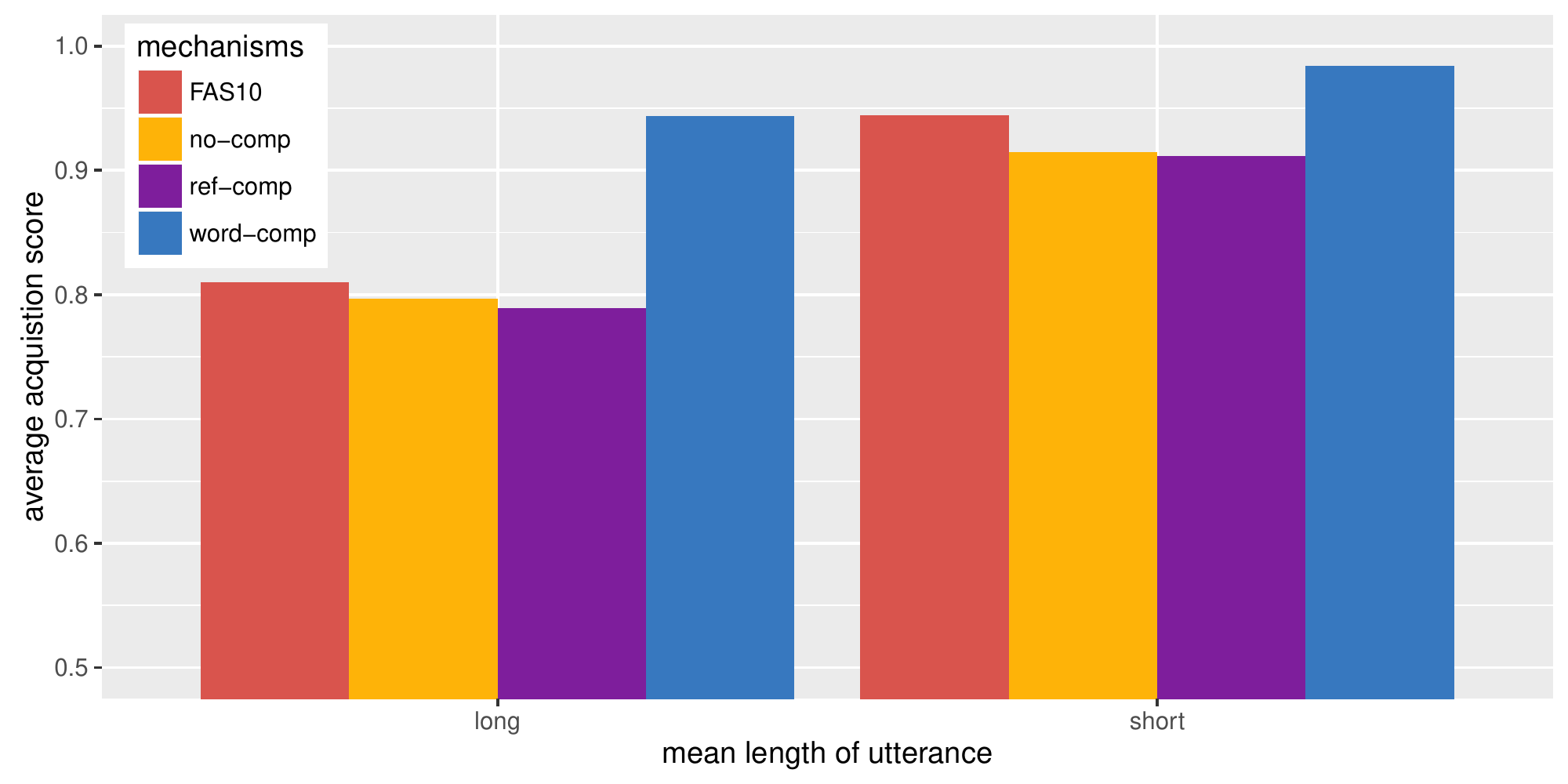}
\caption{Split over mean length of utterance.}
\label{fig:res-acq-mlu}
\end{subfigure}
\begin{subfigure}{.32\textwidth}
\includegraphics[width = \linewidth]{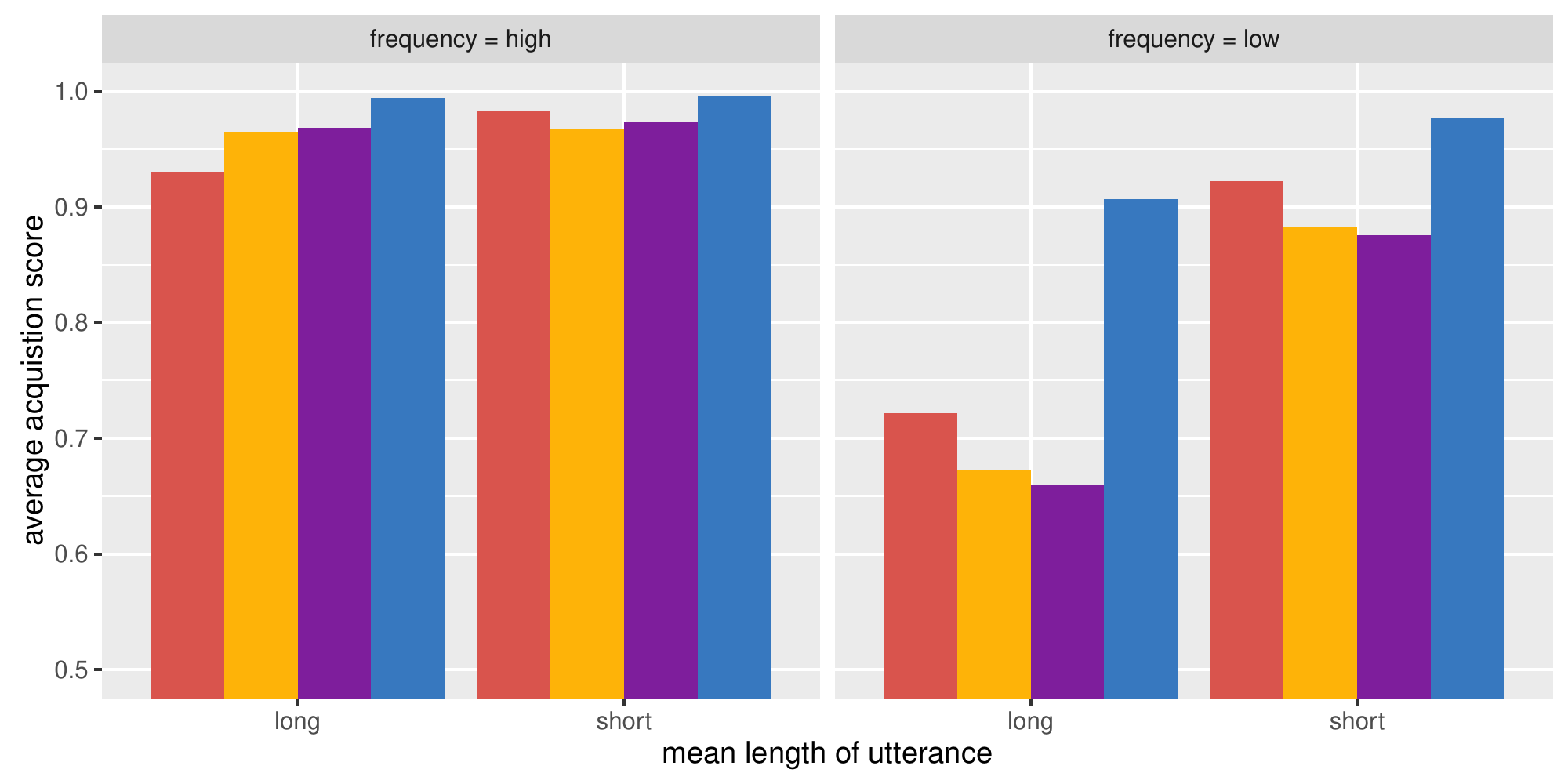}
\caption{Split over MLU and word frequency.}
\label{fig:res-acq-freq-mlu}
\end{subfigure}
\caption{Average acquisition score after 20K input items.}
\vspace{-0.4cm}
\end{figure*}

Over time, all models converge to high average $\mathbf{acq}$ scores
(\Figure{fig:res-acq}) and proportions of words learned
(\Figure{fig:res-proplearned}), but with substantial differences between them.
Notably, we find that on the average $\mathbf{acq}$ score, the word-comp
formulation performs better than the original FAS ($.96$ vs. $.86$), while the
ref-comp and no-comp models do not learn the representations as well (both
$.83$).

Two factors may underlie the varying performance of the models: the semantic
grouping of features into referents (distinguishing our models from FAS), and
the type of in-the-moment competition (and resulting type of mutual
exclusivity).
For the first factor, the word-comp mechanism provides the most direct
comparison to FAS: it uses the same direction of bias -- in which words compete
to align with the elements of the scene -- but using referents instead of
features.  The grouping into referents appears to improve learning.  When
aligning features individually as in FAS, the correct features for a word may
be aligned more or less strongly (depending on competition for each from other
words), so that the overall meaning probability vector may not converge as
easily to the full set of correct features.  By contrast, when a word has a
strong alignment with the correct referent -- which corresponds to the
gold-standard meaning of the word -- \textit{all} features of the referent are
boosted in the meaning probability of the word, yielding improved learning in word-comp over FAS.

Second, we find an interesting asymmetry between the two mechanisms involving
competition between the words for a referent (word-comp) and between the
referents for a word (ref-comp).  Each imposes a conditional probability
formulation of competition, but word-comp performs much better, with ref-comp
behaving no better than the no-comp model.  In fact, the advantage of using
referents instead of individual features is completely eliminated in both the
no-comp and the ref-comp mechanisms, as both perform worse than FAS.  This is
especially surprising given that \citeA{alishahi.etal.2012} used the ref-comp
alignment mechanism in their work that modeled human behaviour in a language
learning task. (We return to this point below.)

The source of this asymmetry, we believe, is the deployment of learned
knowledge by the model. In both the no-comp and the ref-comp model
(\Figure{fig:align}(b), (c)), a learned word meaning is compared to the
referents in isolation from the learned meanings of the other words in the
utterance. In this set-up, the knowledge of other word meanings cannot help to
guide the model to determine how good a word's alignment to some referent is.
By contrast, the word-comp model (\Figure{fig:align}(d)) tunes the alignments
by comparing how similar various learned word meanings are to a referent.

One might expect that mutual exclusivity in the reverse direction (as in the
ref-comp model) would achieve the same effects: Tuning the similarity between a
word meaning and a referent by the similarity between that word meaning and all
other referents should guide the model to correct associations more quickly
than not doing so. However, we do not find this effect. We will return to the
reason for this lack of effect in the section on the role of frequency. 

Competition is clearly important in focusing alignments and facilitating
learning, but only in the context of appropriately constraining information:
the most effective learning occurs when the competition draws on the maximal
amount of learned knowledge in the model, in the form of the developing meaning
probabilities.  In what follows, we consider the impact of increased ambiguity
in forming alignments, or decreased knowledge about words, to see how these
factors impact these various mechanisms.
Because the proportion of words learned shows similar relative behaviours to
the \textbf{acq} score, in the remaining analysis we focus on comparing
\textbf{acq} scores of each of the models after $20$K inputs.

\vspace{.1cm}
\noindent\textbf{The Role of Frequency}

Children are able to learn word meanings in various conditions, sometimes after
only a few observations. Previous research suggests that children use biases
such as mutual exclusivity to guide their learning. Learning low-frequency
words is also a challenge for computational models, and understanding the
mechanisms that improve learning from little evidence can shed light on how
children address this issue.  The type of competition in the various models
under study here plays an important role in their performance on low-frequency
words.  \Figure{fig:res-acq-freq} shows that for the two models with
competition over words -- the FAS and word-comp models -- there is no decrease
in performance for words of low frequency ($<5$) compared to high frequency
($>10$), while for the other two models, no-comp and ref-comp, there is a
dramatic drop off in learning.  

Specifically, the competition among words in the FAS and word-comp models --
which maximizes the use of learned knowledge in focusing alignments -- appears
to play a crucial role in enabling these models to learn low-frequency words.
Comparing the alignments in \Figure{fig:align-word2refs} and
\Figure{fig:align-ref2words} in the face of a novel word and its novel referent
(as an extreme case of low frequency) will clarify the utility of the learned
meaning probabilities.  In the word-comp model (\Figure{fig:align-ref2words}),
the meaning probabilities of previously-seen words competing for a new referent
will not have a very good match to the feature vector for the new referent
(since their probabilities will have been adjusted to better fit referents they
have been seen with).  The novel word will have uniform meaning probabilities
that will enable it to better match the new referent, and thus will have a
stronger alignment than previously-seen words.  By contrast, in the ref-comp
model (\Figure{fig:align-word2refs}), the uniform probabilities of the new word
will equally match all the referents competing for it, whether they have been
seen before or not.  There is no prior knowledge in the model in this
competition that indicates the previously-seen referents have a better fit with
other words.  Thus a competition among words works well for novel or
low-frequency words by drawing on the fact that previously-seen words will not
compete as strongly for a new(er) referent.  In short: a new word can in
principle go equally well with any referent in the situation, but a new
referent not equally well with any word in the utterance.

\vspace{.1cm}
\noindent\textbf{The Role of Utterance Length}

Above, we found that the different types of competition gave more pronounced
results for low-frequency words than for high-frequency ones.  Similarly, we
can explore whether there is a differential impact of utterance length on the
different models. 
To simulate this, we manipulated the input generation procedure so that the
model was trained only on utterances of length $5$ or higher (long-corpus), or
$3$ and lower (short-corpus). 
Looking at \Figure{fig:res-acq-mlu}, we observe that the acquisition scores are
globally lower when the models are trained on long sentences only, likely due
to the fact that there is more uncertainty about which words and which
referents belong together.

Here we see that the word-comp model is the only one to not substantially
decline in performance when comparing learning on the short-corpus and
long-corpus.  
While the competition over words seems to help the FAS and word-comp models
equally in dealing with low-frequency words, here the bundling of features into
referents as in word-comp is also necessary for performance to be robust to the
added ambiguity of long utterances.  The FAS model cannot ``scale up'' to deal
with the very long unstructured lists of features in the long-corpus input.  We
can also now suggest why the \citeA{alishahi.etal.2012} model (the ref-comp
approach) worked well in their experiments but not here: the utterances they
used all had two words, unlike the naturalistic data we train on above,
indicating that ref-comp also cannot scale effectively.
Interestingly, as shown in \Figure{fig:res-acq-freq-mlu}, we see that the
word-comp model is particularly robust to the challenge of learning
low-frequency words in the corpus of longer utterances, with a very small
decrease in performance compared to the other models.

\vspace{.1cm}
\noindent\textbf{The Role of Referential Uncertainty}

\begin{figure}
\centering
\includegraphics[width = 0.667\linewidth]{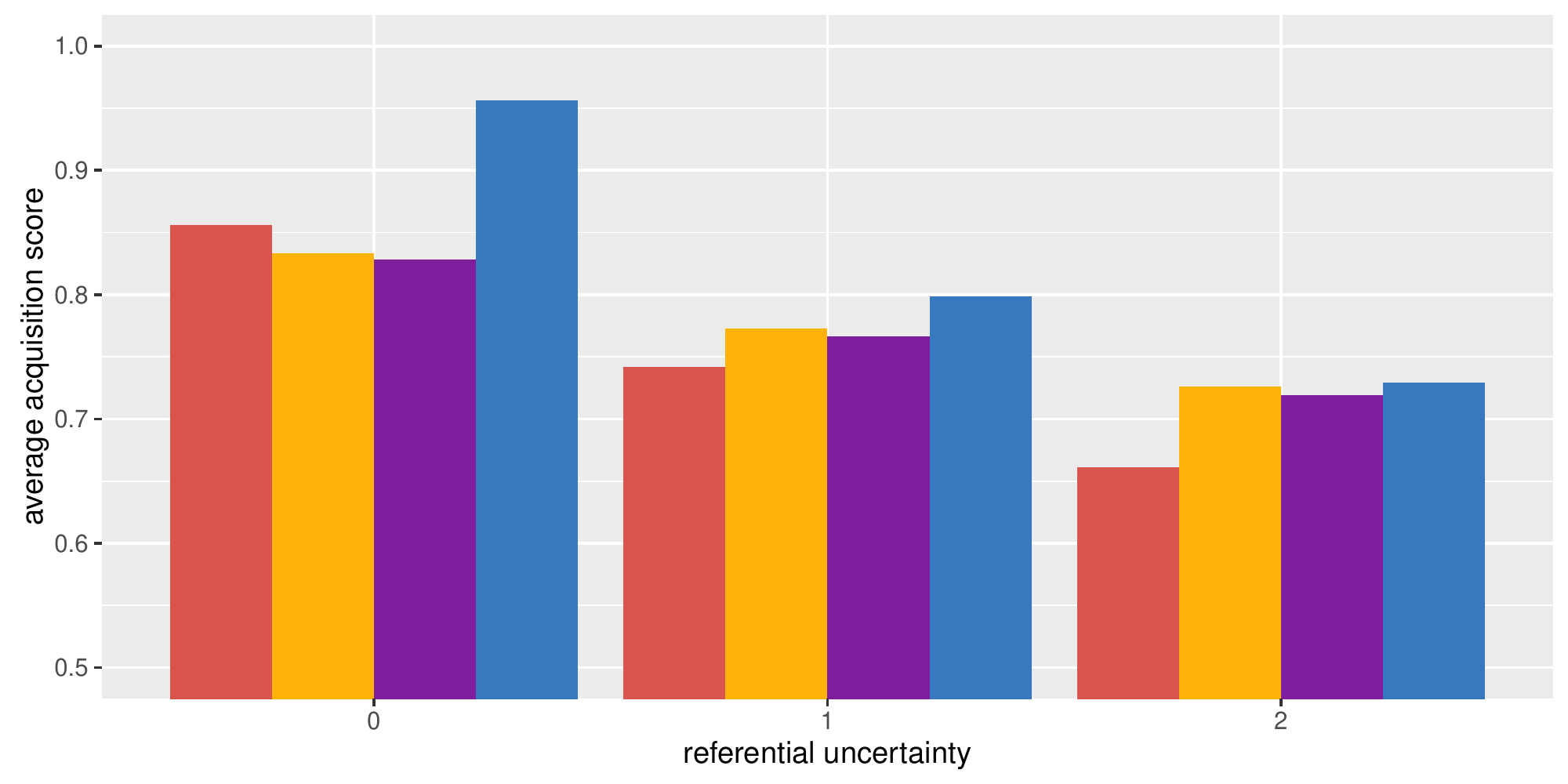}
\vspace{-0.2cm}
\caption{Average acquisition score after 20K input items, split over different amounts of referential uncertainty.}
\vspace{-0.5cm}
\label{fig:res-acq-ru}
\end{figure}

To explore the impact of referential uncertainty -- the occurrence of many more
potential referents in a scene than there are words -- we create a subcorpus
that uses every $i^{th}$ utterance from our full corpus, and uses the
utterances in between those to generate ``extra'' referents in the scenes for
utterances in the subcorpus.
Here we report results on $20$K inputs with referents added to each scene $S_i$
from $0$, $1$, or $2$ utterances in addition to referents taken from utterance
$U_i$.
\Figure{fig:res-acq-ru} presents the results for no referential uncertainty,
along with the two added levels of uncertainty.  As we expect, the learning
performance of all models degrades with higher referential uncertainty.
However, in contrast to our previous results, here there is little benefit from
either word-based competition or feature bundling.  The high degree of
ambiguity introduced by these levels of referential uncertainty may be better
dealt with by attentional mechanisms that focus joint attention on a likely
subset of relevant referents prior to alignment.

\section{Conclusions and Future Work}

Previous research shows that children are sensitive to the cross-situational
statistics of their environment: i.e., they can use the regularities across
different situations to learn word meanings.
However, the detailed mechanisms responsible for cross-situational word
learning are still not fully understood, such as precisely what information is
used from each observation in identifying the correct word meaning, and how
this information is incorporated in the accumulated knowledge about a word.
Moreover, children are good at learning word meanings in a variety of
situations: they can learn a novel word from a few example and also acquire
words from ambiguous/noisy conditions. 
Previous research has suggested that children are equipped with biases that
guide them in word learning by reducing the difficulty/ambiguity of a learning
situation. 
The necessity of these biases in children, and whether they are innate or
learnable, are issues that have been debated among cognitive scientists. 

Here, we show that one such bias -- the mutual exclusivity bias that limits the
number of meanings a word takes -- can be modeled as a competition mechanism
during in-the-moment learning. The competition exists when the model assesses
possible word and referent associations with conditional probabilities as
opposed to counts. In other words, the bias or competition is a learning
mechanism that is able to condition in-the-moment learning to the learned
knowledge of word meanings. We observe that the role of the bias is
particularly significant when the learning is more challenging: for example,
for learning low-frequency words or from longer utterances.
Previous research has investigated how cognitive processes such as memory and
attention interact with cross-situational word learning
\cite<\eg,>{nematzadeh.etal.2012.cmcl}. Future work should study how these
cognitive processes affect the in-moment-learning.

\setlength{\bibleftmargin}{.125in}
\setlength{\bibindent}{-\bibleftmargin}
\bibliographystyle{apacite}
\bibliography{nematzadeh}

\end{document}